\documentclass[runningheads]{llncs}
\usepackage{graphicx}
\usepackage{times}
\usepackage{soul}
\usepackage{url}
\usepackage{graphicx}
\usepackage{amsmath}
\usepackage{booktabs}
\usepackage{subfigure}
\usepackage{subfig}
\usepackage{url}
\usepackage{array}
\usepackage{amsmath}
\renewcommand\tabcolsep{8.0pt}

\begin{document}
\title{Transformer-based Language Model Fine-tuning Methods for COVID-19 Fake News Detection}
\titlerunning{Language Model Fine-tuning Methods for COVID-19 Fake News Detection}
% Bin Chen\thanks{contribute equally with first author.} \and
%
\author{Ben Chen \and
Bin Chen \and
Dehong Gao \and
Qijin Chen \and
Chengfu Huo \and \\
Xiaonan Meng \and
Weijun Ren \and
Yang Zhou
}
% Ben Chen, Bin Chen, Dehong Gao, Qijin Chen, Chengfu Huo, Xiaonan Meng, Weijun Ren, Yang Zhou
\authorrunning{Ben Chen et al.}
\institute{Alibaba Group, Hangzhou, China \\
\email{\{chenben.cb,cb242829,dehong.gdh,qijin.cqj,\\
chengfu.hcf,xiaonan.mengxn, afei, yngzhou\}\\
@alibaba-inc.com}}
\maketitle              % typeset the header of the contribution

% \footnotetext[2]{Corresponding authors.}

\begin{abstract}
With the pandemic of COVID-19, relevant fake news is spreading all over the sky throughout the social media. Believing in them without discrimination can cause great trouble to people's life. However, universal language models may perform weakly in these fake news detection for lack of large-scale annotated data and sufficient semantic understanding of domain-specific knowledge. While the model trained on corresponding corpora is also mediocre for insufficient learning. In this paper, we propose a novel transformer-based language model fine-tuning approach for these fake news detection. First, the token vocabulary of individual model is expanded for the actual semantics of professional phrases. Second, we adapt the heated-up softmax loss to distinguish the hard-mining samples, which are common for fake news because of the disambiguation of short text. Then, we involve adversarial training to improve the model's robustness. Last, the predicted features extracted by universal language model RoBERTa and domain-specific model CT-BERT are fused by one multiple layer perception to integrate fine-grained and high-level specific representations. Quantitative experimental results evaluated on existing COVID-19 fake news dataset show its superior performances compared to the state-of-the-art methods among various evaluation metrics. Furthermore, the best weighted average F1 score achieves 99.02$\%$. 

\end{abstract}
\section{Introduction}
The development of social media, such as Twitter and MicroBlog, has greatly facilitated people's lives. We can get real-time news from almost anywhere in the world. However, fabrications, satires, and hoaxes mixed in real reports often mislead people's judgments, especially during the pandemic. For example, "CDC Recommends Mothers Stop Breastfeeding To Boost Vaccine Efficacy"\footnote{https://www.snopes.com/fact-check/breast-practices/} and "Consuming alcohol beverages or vodka will reduce risk of COVID-19 infection"\footnote{https://www.usatoday.com/story/news/factcheck/2020/03/20/} are two common rumors during the epidemic and caused panic among the masses. Therefore, fake news detection is necessary and we hope to design an effective detector which could quickly distinguish whether the news is fake or not according to its title or summary. It is usually formulated as the sequence classification problem in general.

Text classification is a fundamental task in natural language processing (NLP), and transformer-based language models have achieved excellent performance in general domains thanks to large corresponding corpora and fined-designed pre-training skills (MLM/NSP/SOP) \cite{BERT,RoBERTaa,ALBERT}. However, they usually perform weak for specific domain. One main reason is the professional phrases are rare in general corpora and existing tokenizers (e.g. byte-pair-encoding, wordpiece, and sentencepiece\footnote{https://github.com/huggingface/tokenizers}) would split them into many sub-tokens, and this operation hampers their actual semantics. Even if data of specific domain is collected and used for down-stream fine-tuning, the limited token vocabulary also fails to get the full meaning. Recently, \cite{ref_article_covid-19,FakeCovid-A,ref_article_cd_BERT} have devoted to collected amounts of COVID-19 data. Especially for \cite{ref_article_cd_BERT}, it also further trains one transformer-based model named CT-BERT with part of these annotated data, making a 10-30$\%$ marginal improvement compared to its base model on classification, question-answering and chatbots tasks related to COVID-19. But for the specific fake news detection, its insufficient learning of limited corpus contents and incomplete hard samples mining make it hard to achieves one impressive result. Furthermore, excessive further-training weakens the model's understanding of common sense, resulting in some incomprehensible mistakes. % 最好找一个例子

In these paper, we try to optimize the performance of transformed-based language models for COVID-19 fake news detection. For individual model, firstly the token vocabulary is expanded with most frequent professional phrases for getting the actual semantics without no split. Second, we adapt the heated-up softmax loss to distinguish the hard-mining samples, which are common for fake news because of the disambiguation of short text. Then, adversarial training \cite{FreeLB} is involved to improve the model's generalization and robustness. Last, the predicted features extracted by universal language model RoBERTa \cite{RoBERTaa} and domain-specific model CT-BERT \cite{ref_article_cd_BERT} are fused by one multiple layer perception to integrate fine-grained universal and high-level specific representations. Quantitative experimental results evaluated on existing COVID-19 fake news dataset \cite{ref_article_covid-19} show these methods superior performances compared to the state-of-the-art methods among various evaluation metrics. Furthermore, the best weighted average F1 score achieves 99.02$\%$.

%社交媒体，例如Twitter和微博的发展极大地便利了人们的生活。通过新闻快讯，我们可以实时、快速地获取到世界任何地方发生的任何事情。然而，混杂在真实报道中的fabrication, satire and haxes 可能会误导人们的判断，尤其是在疫情期间。 

%1) 社交媒体信息的碎片化，简单而且没有完整的语义信息
%2）短文本不能为大型的BERT-based模型提供完整的上下文语义理解

\section { Related Work}
% \subsection {Dealing with hard mining example }
\subsection {Text classification task with adversarial training methods}
Adversarial training is firstly designed to increase the model robustness through adding small perturbations to training data, but it also increases the generalization ultimately \cite{FreeLB}. In the field of computer vision, mainstream gradient-based attack \cite{ref_article1}, optimization-based attack \cite{ref_article2} and generation-based attack \cite{ref_article3} have achieved impressive results. In recent years, more and more adversarial training tips \cite{ref_article_NLPattack,ref_article_NLPattack2,FreeLB} have been proposed for natural language processing tasks. Different from computer vision, where the image pixel is continuous in the fixed space, so it is suitable to add noise perturbation based on gradient method. However, in natural language processing tasks, the input is usually a discrete text sequence, so it is impossible to realize the antagonistic training by adding disturbance to the one-hot layer of the input text. From the Goodfellow's work \cite{ref_article_NLPattack}, they realized the adversarial training by adding embedding disturbance into the embedding layer. Usually in the CV task, according to empirical conclusions, the adversarial training tends to make the performance of the model on the normal samples worse, while in the NLP task, the work of \cite{ref_article_NLPattack} shows that the generalization ability of the model is stronger. In this paper, we use the method of adversarial training to improve the generalization ability of the model in the task of Fake New detection.

\subsection {Model fusion approaches for text classification }
Traditional machine learning models have proved that ensemble learning play an important roles in improving the model effect, such as Bagging and Boosting. The main reason lies in the complementary feature among models helps the model to make correct judgment on the results. In recent years, a series of model fusion methods have also appeared in the field of deep learning. The methods for model fusion mainly include feature-based methods\cite{ref_article_fusion_feature} and score-based methods\cite{ref_article_fusion_score}. Feature level fusion method is suitable for models of different categories, such as CNN and BiGRU, and can extract word-level features and syntax-level features simultaneously\cite{ref_article_fusion_feature}. While the fusion methods based on the score level are more applicable to similar structure for models, which obtain the final result by voting. This paper combines CT-BERT and RoBERTa by using the fusion method at the score level, while ensuring the universality of the model and its professionalism simultaneously in the task of Fake News Detection.
\subsection {Fake news detection }
Fake news is intentionally written to mislead readers to believe false information, which makes it difficult and nontrivial to detect based on news content. Most existing methods identify false news by combining richer data, which provide a data repository that include not only news contents and social contents, but also dynamic information\cite{FakeCovid-A}. However, such methods lack universality and cannot play a role in new fields. Hence, semi-supervised and unsupervised methods are proposed\cite{ref_article_semi}, which try to make a trade-off between the amount of training data and the final training accuracy. Based on the excellent representational ability of deep pre-trained model (E.g, BERT, ALBERT), our method tries to get a well result by utilizing a small amount of data from domain special fields.

\section {Methodology}
\subsection {Problem Definition}
As described above, in this paper we convert COVID-19 fake news detection as one typical single sentence classification, which means the proposed detector can judge whether one news sentence is true or false according to its semantic meaning. It can be expressed formally as: giving a news sentence $x=t_1,t_2,t_3,t_4,t_5,....$, the detector should predict one correct label $y \subseteq {\{0,1 \}}$. And so the corresponding optimization goal is to learn the $\theta $ and maximize the loss function $L(y|x,\theta)$.

\subsection { Our proposed network}
As shown in Fig.1 is the structure of our proposed network. It is derived from the most famous language model $-$ BERT, and we involve some practical methods to enhance the ultimate performance from various perspectives. Below we will introduce each module in detail.

\subsubsection {Training with additional tokens}
For specific domain text, there are many professional phrases and existing tokenizers will split them into many sub-tokens, resulting in a misunderstanding of their actual meanings. In this paper, we count 6 most frequent tokens in train and validation data which will be split with original method and add them in the existing token vocabulary of CT-BERT. There are:

%\begin{itemize}
\vspace{5pt}
$\bullet$ \; \; covid$-$19, covid19, coronavirus, pandemic, indiafightscorona, lockdown.
\vspace{5pt}
%\end{itemize}

Subsequent ablation experiments will prove the effectiveness of this method.
\vspace{-1pt}

\begin{figure}[hbt]
  \begin{center}
\subfigure[]{
  \centering 
\resizebox*{6.5cm}{3cm}{\includegraphics{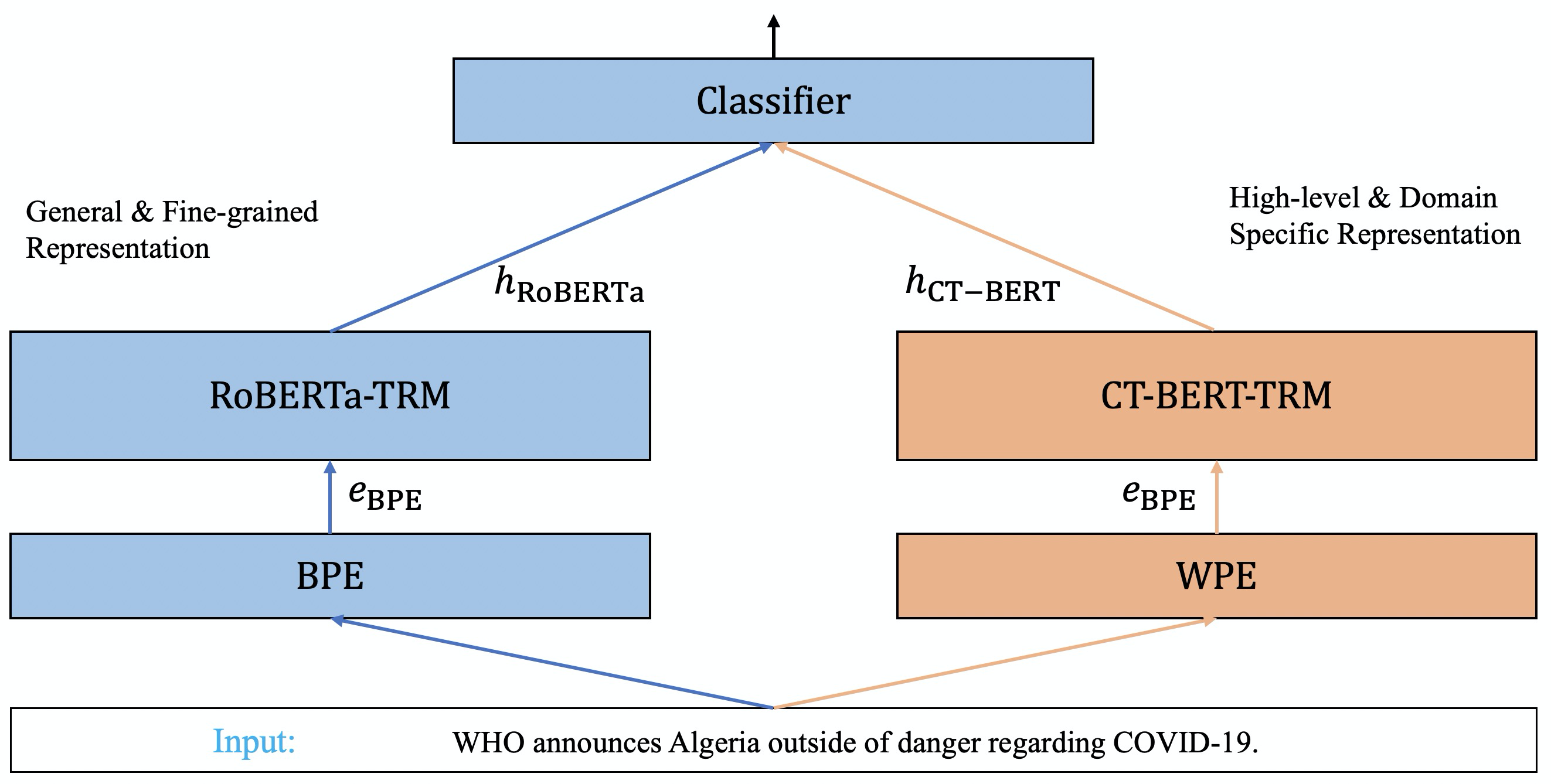}}}
 \subfigure[]{
   \centering 
\resizebox*{9.2cm}{6cm}{\includegraphics{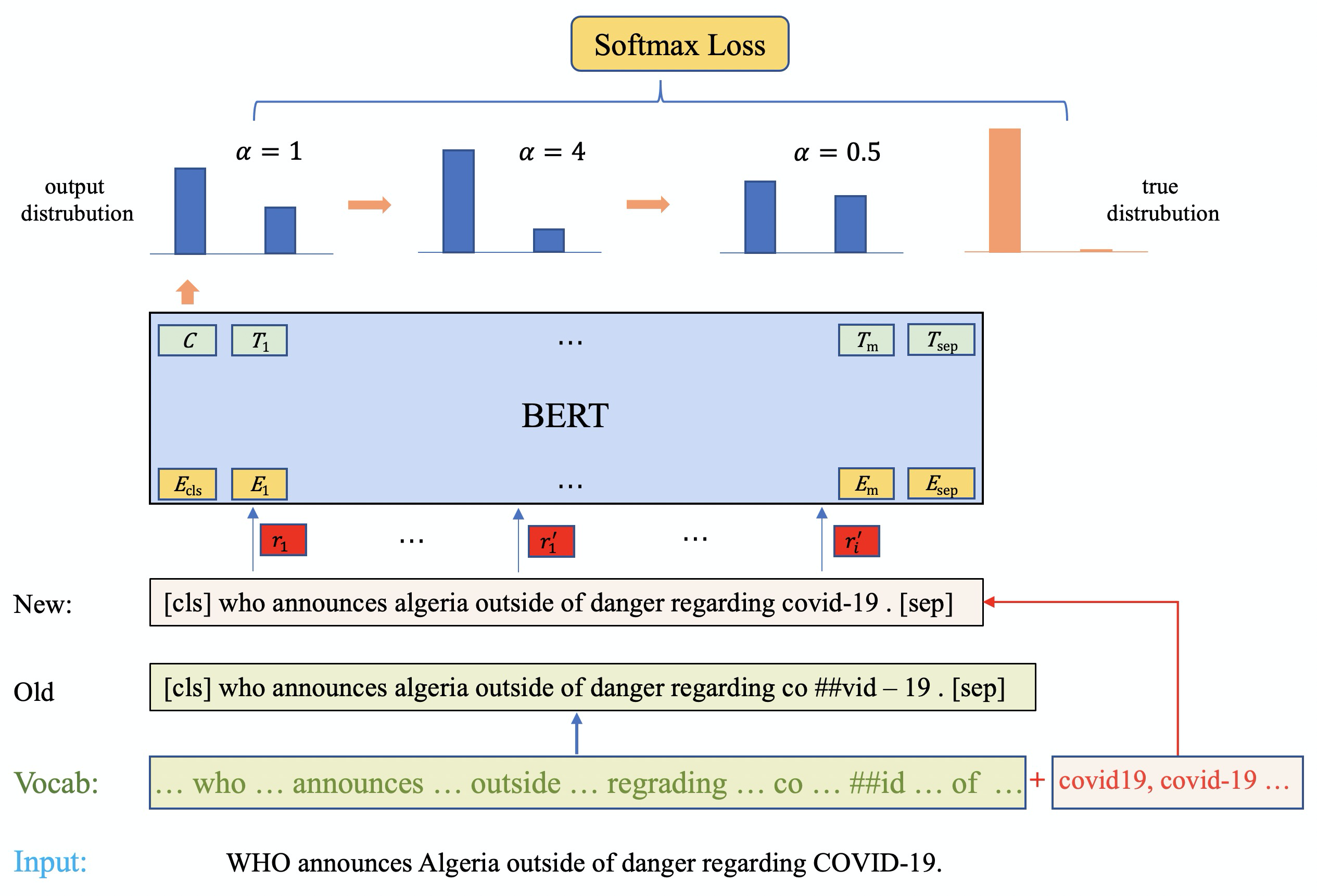}}}\hspace{5pt}
  \vspace{-10pt}
  \caption{The framework of our proposed method. (a) is the overall framework of the fused models and (b) is three improved modules added to each model. The bottom part shows the result of adding new tokens to the existing vocabulary. The middle red boxes represent the gradient perturbations added to the word embedding. And the top blue histograms exhibit how the heated-up softmax affects the distribution.} 
  \vspace{-25pt}
 \label{figure3}
\end{center}
\end{figure}
%\vspace{-10pt}
% \begin{figure}[ht]
%   \centering 
%   \includegraphics[width=4.5in]{image/main_image4.png}
%   \caption{The framework of our proposed network.The red boxes represents the gradient perturbations added to the word embedding. The blue dotted line implies that the model pays more attention to the specific word to distinguish the hard mining examples because of the heated-up softmax existed.} 
% \end{figure}

\subsubsection {Optimization with heated-up softmax loss function}
For binary classification task, we utilize the cross entropy loss as our loss function. In order to remind model to pay more attention to the hard mining examples, we introduce the heated-up softmax to replace the origin activation  function\cite{Heated-Up}. Initially, the heated-up softmax is expressed as follow:
\begin{equation}
p(m|x) = \frac{exp(z_{m}/T)}{\sum_{j=1}^{M}exp(z_{j}/T)}=\frac{exp(\alpha z_{m})}{\sum_{j=1}^{M}exp(\alpha z_m)}
\end{equation}
where $\alpha$ denotes the temperature parameter. As can be seen from Eq.1, the larger the parameter $\alpha$ is, the greater the gradient value is for the hard mining samples, and the model pays more attention to the hard mining samples. The smaller the $\alpha$ value, the smaller the attention to hard mining sample, boundary sample and easy sample. Therefore, at the beginning of the training, we first set a large $\alpha$ and let the model focus on learning the difficult sample, and then reduce the $\alpha$. The model began to pay attention to the boundary sample, and when the hard mining sample and boundary sample were almost completed, then further reduced $\alpha$ to fine tuning. As depicted in Fig.1(b), the factor $\alpha$ affect the distribution of the last output layer, which is efficient for classify the hard example to adjust the $\alpha$ as training going.

\subsubsection {Gradient-based adversarial training }
Adversarial training \cite{ref_article_goodfflow_adv} is a novel regularization method for classifiers to improve model robustness for small, approximately worst case perturbations. The loss function we used with adversarial training unit can be formulated as:
\begin{equation}
    -log \; p(y|x+r_{adv};\theta) \; \; \;where  \;  \; r_{adv} =argmin_{r,||r||<=\epsilon} \;log \; p( x|x+r;\hat{\theta})
\end{equation}
where $r_{adv}$ denotes the perturbations , $ \hat{\theta}$ denotes the parameters of current network, and $ \theta$ denotes the  parameters after one-step gradient optimization of the network. We can find from the Eq.2 that our core objective is to add a small perturbation $r$ which could disable the current classifier, then optimize the classifier, and optimize the network to maximize the model's ability to correctly classify samples. However, we cannot calculate this value exactly in general, because exact minimization with respect to $r$ is intractable for many interesting models such as neural networks.
Goodfellow et al. \cite{{ref_article2}} proposed to approximate this value by linearizing $log \; p(y | x; \hat{\theta})$ around $x$. With a linear approximation and a L2 norm constraint in Eq.(2), the final adversarial perturbation is

\begin{equation}
r_{adv} = - \epsilon \frac{g}{||g||_{2}} \; \; where \; \; g=\nabla_{x} log p(y|x;\hat{\theta})
\end{equation}

This perturbation can be easily computed using backpropagation in neural networks.

In Fig.1(b), the red box represents the perturbation calculated by Eq.3. perturbation is added to the embedding layer, and the robustness of the model is enhanced by adversarial training.

\section {Experimental Results}
In this section, we will evaluate performance of our method in Fake News Detection task. We compared our approach to several pre-trained models that perform well on Glue tasks\footnote{https://gluebenchmark.com/}, including BERT, ALBERT, and RoBERTa, and each model included the basic version (X-Base) and the Large version (X-Large).

In order to reflect the importance of each module in our model,  we also perform the ablation studies on our experiments. Specifically, we divide our approach into 3 different methods and test the performance of each module: COVID-Twitter-BERT (benchmark model, refer to as CT-BERT model below) \cite{ref_article_cd_BERT}, CT-BERT-FGM(including adversarial training module with fast gradient method),CT-BERT-HL (including heated-up softmax module),CT-BERT-New-Tokens (including new tokens training ), Ro-CT-BERT (ours).

\subsection {Experimental setting}
The data we used for training and evaluation is the online-collected COVID-19 fake news dataset \cite{ref_article_covid-19}. The sources of them are various social-media platforms such as Twitter, Facebook, Instagram, etc. It contains 6420 / 2140 / 2140 raw news sentence for training / validation / test. The real news sentence is as follow:
$$Wearing \; mask \; can \; protect \; you \; from \; the \; virus.$$

While the fake one is shown as follow:
$$If \; you \; take \; Crocin \; thrice \; a \; day \; you \; are \; safe.$$

As a side note, for the data preprocessing, we follow the baselines in \cite{ref_article_covid-19} to remove all links, non alphanumeric characters (e.g. unicode emotions) and English stop words, which all would bring great interference to the effective detection. In order to train the more distinguishable models, After each evaluation we will reserve the mis-classified samples in training and validation set and replace some (1$\sim$2) words with their synonyms or remove these words directly. The extended data will be added to the training set for the next round of training.

To evaluate the performance of different methods, three popular metrics are adopted, namely weighted Precision, weighted Recall and weighted F1. The definitions are as follows:

\begin{equation}
     Precision_{weighted} = \frac{\sum_{i=1}^{n}Precision_i*w_i}{n}
\end{equation}

\begin{equation}
     Recall_{weighted} = \frac{\sum_{i=1}^{n}Recall_i*w_i}{n}
\end{equation}

\begin{equation}
     F1_{weighted} = \frac{2 * Precision_{weighted} * Recall_{weighted}}{Precision_{weighted} + Recall_{weighted}}
\end{equation}

\vspace{5pt}
where $n$ represents the number of classes, $w_i$ represents the radio of true instances for each label.

Our implementation is based on the online available natural language library \textbf{Transformers}\footnote{https://github.com/huggingface/transformers}. We adopt the initial learning rate of 2e-5 with warm-up rate of 0.1. The batch size is selected as 64 for training and 128 for validation and test. The temperature parameter $\alpha$ is set as 4 in first 10 epochs, as 1 in  middle 10 epochs and as 0.5 for last 10 epochs. Each sequence length is limited to 128 tokens. All experiments are performed using PyTorch on a Telsa V100 GPU with the optimizer selected as Adam.

\subsection {Performance in Fake News Detection in English }
In this paper, we investigate the most cutting-edge models to tackle fake New Detection tasks, including the highly versatile BERT \cite{BERT}, ALBERT \cite{ALBERT}, and RoBERTa \cite{RoBERTaa} and their large versions; In addition, we investigate COVID-Twitter-BERT (CT-BERT), which is trained on a large corpus of Twitter messages on the topic of COVID-19. We utilize the pre-trained model files provided on the official website to initialize corresponding parameters, and then fine tuning on the COVID-19 dataset \cite{ref_article_covid-19}. Each model used the best result as the final experimental result. For the sake of description, we call our model as \textbf{Ro}bust-\textbf{C}OVID-\textbf{T}witter-BERT (\textbf{Ro-CT-BERT}). The experimental results are shown in Table 1.

\begin{table}
\centering
\vspace{-15pt}
\caption{Experimental comparison results on fake news detection task.}\label{tab2}
\vspace{7pt}
\begin{tabular}{|c|c|c|c|c|}
\hline
Method & Accuracy &Precision&Recall& F1 \\
\hline
BERT-base &  0.978505& 0.978574& 0.978505	& 0.978497\\
BERT-large &  0.980374	& 0.980407& 0.980374&	0.980369\\
RoBERTa-base & 0.983645& 0.983755& 0.983644& 0.983638\\
RoBERTa-large & 0.985981&	0.986081&	0.985981	& 0.985976\\
ALBERT-base& 0.973365& 0.973419& 0.973365& 0.973356\\
ALBERT-large& 0.973832&0.973897&0.973832&0.973823\\
ALBERT-xlarge& 0.974299& 0.974665& 0.974299& 0.974276\\
CT-BERT&0.984112&0.984161&0.984112&0.984115\\
Ro-CT-BERT  & \textbf{0.990187} & \textbf{0.990218}& \textbf{0.990187}& \textbf{0.990185} \\
\hline
\end{tabular}
\end{table}

\vspace{-5pt}
From the experimental results, We can see that Ro-CT-BERT can get superior performance than state-of-the-art methods on the metrics weighted average accuracy, precision, recall and F1 score. Universal language models BERT, ALBERT get all metric value lower than 0.981, while RoBERTa is much better than them because of fine-designed key hyperparameters and larger training data size.
Although the CT-BERT model has been trained in a large number of Messages on the Topic of COVID-19, our model's F1 score is still 0.006 points higher than it's F1 score. This promotion is very difficult because we need to make the correct classification for hard mining samples. We attribute the improvement of the model to the hard mining samples learning and effective fusion of fine-grained and high-level representations. In the next section, we will demonstrate the effect of each module on the model through ablation experiments.

\subsection {Ablation studies for Ro-CT-BERT}
In order to verify the effectiveness of the improved modules involved in Ro-CT-BERT, we also conduct several ablation experiments. We mainly compare the influence of three modules on the model, which called adversarial training, heated-up softmax loss function and addition of new Token. For fairly comparison, we take CT-BERT as the benchmark model and add three modules for subsequent experiments, respectively. These three models are successively referred as attack-training-CT-BERT (CT-BERT-FGM), heated-up softmax loss CT-BERT (CT-BERT-HL) and new-taken CT-BERT (CT-BERT-New-Tokens). The model with all three modules is referred as three-modules-CT-BERT (CT-BERT-TRM). Results of ablation experiments are shown in Table 2.

\begin{table}
\centering
\vspace{-15pt}
\caption{Experimental comparison results on Fake News Detection task.}\label{tab1}
\vspace{7pt}
\begin{tabular}{|c|c|c|c|c|}
\hline
Method & Accuracy &Precision&Recall& F1 \\
\hline
CT-BERT&0.984112&0.984161&0.984112&0.984115\\
CT-BERT-FGM& 0.986449&0.986451 &0.986449 &0.986448\\
CT-BERT-HL  &0.986916 & 0.986971 & 0.986916 & 0.986912\\
CT-BERT-New-Tokens&0.984579	& 0.984623&	 0.984579& 0.984575\\
CT-BERT-TRM&0.987851	& 0.987888&	 0.987851& 0.987848\\
Ro-CT-BERT  & \textbf{0.990187} & \textbf{0.990218}& \textbf{0.990187}& \textbf{0.990185} \\ \hline
\end{tabular}
\vspace{-5pt}
\end{table}

It can be seen that compared with the TD-BERT model without any other tricks, the classification effect of the other three models is improved to a certain extent, especially the heated-up softmax loss, which increases the generalization ability of the model and has a strong classification ability for the hard mining samples. Furthermore, these three modules combined produces one better result. Lastly, Ro-CT-BERT fuse the predicted features of CT-BERT-TRM and RoBERTa-TRM (three-module RoBERTa) and get the highest score, indicating that the integration of fine-grained and high-level specific representations helps to understand the text semantics more comprehensively.

\section{Conclusions}
In this work, we propose a transformer-based Language Model Fine-tuning approach for COVID-19 Fake News Detection\cite{CONSTRAINT-2021}. The length of adopted news sentences for this task is short, and lots of professional phrases are rare in common corpora. These two distinct features make universal and specific language models all fail to make a correct distinction whether  news is fake or real. To address these problems, we respectively introduce new tokens for the specific model vocabulary for better understanding of professional phrases, model adversarial training to improve the robustness, and heated-up softmax loss function to distinguish the hard-mining sample. Lastly, we also fuse the predicted features extracted by universal language model and domain-specific model to integrate fine-grained and high-level specific representations. These methods are verified to be useful for improving the transformer-based model's ability, and finally, our approach achieves super performance compare with state-of-the-art methods on the COVID-19 fake news detection.


\begin{thebibliography}{8}
\bibitem{ref_article1}
Carlini, Nicholas, and David Wagner. "Towards evaluating the robustness of neural networks." 2017 ieee symposium on security and privacy (sp). IEEE, 2017.

\bibitem{ref_article2}
Goodfellow, Ian J., Jonathon Shlens, and Christian Szegedy. "Explaining and harnessing adversarial examples." arXiv preprint arXiv:1412.6572 (2014).

\bibitem{ref_article3}
Chaowei Xiao, Bo Li, Jun-Yan Zhu, Warren He, Mingyan Liu, Dawn Song:
Generating Adversarial Examples with Adversarial Networks. CoRR abs/1801.02610 (2018)
a service of Schloss Dagstuhl - Leibniz Center for Informatics	homebrowsesearchabout

% Rubin, V. L. and Vashchilko, T.. Identification of
% truth and deception in text: Application of vector space
% model to rhetorical structure theory. In Proceedings of
% the Workshop on Computational Approaches to Deception Detection, pages 97–106. Association for Computational Linguistics (2012)

\bibitem{ref_article_NLPattack}
Miyato, Takeru, Andrew M. Dai, and Ian Goodfellow. "Adversarial training methods for semi-supervised text classification." arXiv preprint arXiv:1605.07725 (2016).

\bibitem{ref_article_NLPattack2}
Wang, Wenqi, et al. "Towards a Robust Deep Neural Network in Texts: A Survey." arXiv preprint arXiv:1902.07285 (2019).

\bibitem{FreeLB}
Chen Zhu, Yu Cheng, Zhe Gan, Siqi Sun, Tom Goldstein, Jingjing Liu:
FreeLB: Enhanced Adversarial Training for Natural Language Understanding. ICLR 2020

\bibitem{ref_article_fusion_feature}
Xie, Jinbao, et al. "Chinese text classification based on attention mechanism and feature-enhanced fusion neural network." Computing 102.3 (2020): 683-700.

\bibitem{ref_article_fusion_score}
Bhushan S N B, Danti A. Classification of text documents based on score level fusion approach[J]. Pattern Recognition Letters, 2017, 94: 118-126.

\bibitem{ref_article_semi}
Bhattacharjee, Sreyasee Das, Ashit Talukder, and Bala Venkatram Balantrapu. "Active learning based news veracity detection with feature weighting and deep-shallow fusion." 2017 IEEE International Conference on Big Data (Big Data). IEEE, 2017.

\bibitem{Heated-Up}
Xu Zhang, Felix X. Yu, Svebor Karaman, Wei Zhang, Shih-Fu Chang:
Heated-Up Softmax Embedding. CoRR abs/1809.04157 (2018)

\bibitem{ref_article_goodfflow_adv}
Goodfellow, Ian J., Jonathon Shlens, and Christian Szegedy. "Explaining and harnessing adversarial examples." arXiv preprint arXiv:1412.6572 (2014).

\bibitem{ref_article_covid-19}
Patwa, Parth, et al. "Fighting an Infodemic: COVID-19 Fake News Dataset." arXiv preprint arXiv:2011.03327 (2020). 

%\bibitem{ref_article_Hindi_Dataset}
%Bhardwaj, Mohit, et al. "Hostility Detection Dataset in Hindi." arXiv preprint arXiv:2011.03588 (2020).
\bibitem{BERT}
Devlin J, Chang M W, Lee K, et al. BERT: Pre-training of deep bidirectional transformers for language understanding[J]. arXiv preprint arXiv:1810.04805, 2018.
\bibitem{ALBERT}
Lan, Zhenzhong, et al. "ALBERT: A lite BERT for self-supervised learning of language representations." arXiv preprint arXiv:1909.11942 (2019).
\bibitem{RoBERTaa}
Liu, Yinhan, et al. "RoBERTa: A robustly optimized BERT pretraining approach." arXiv preprint arXiv:1907.11692 (2019).

\bibitem{ref_article_cd_BERT}
Müller, Martin, Marcel Salathé, and Per E. Kummervold. "COVID-Twitter-BERT: A Natural Language Processing Model to Analyse COVID-19 Content on Twitter." arXiv preprint arXiv:2005.07503 (2020).

\bibitem{Howt_to_fine}
Sun, Chi , et al. "How to Fine-Tune BERT for Text Classification?." China National Conference on Chinese Computational Linguistics Springer, Cham, 2019.

\bibitem{FakeCovid-A}	Gautam Kishore Shahi, Durgesh Nandini:
FakeCovid-A Multilingual Cross-domain Fact Check News Dataset for COVID-19. CoRR abs/2006.11343 (2020)

\bibitem{CONSTRAINT-2021}	Parth Patwa and Mohit Bhardwaj, et al.
Overview of CONSTRAINT 2021 Shared Tasks: Detecting English COVID-19 Fake News and Hindi Hostile Posts. Proceedings of the First Workshop on Combating Online Hostile Posts in Regional Languages during Emergency Situation ({CONSTRAINT-2021})

\end{thebibliography}
\end{document}